\documentclass[letterpaper, 10 pt, conference]{ieeeconf}  

\IEEEoverridecommandlockouts                              

\makeatletter
\let\NAT@parse\undefined
\makeatother

\usepackage{graphics} 
\usepackage{epsfig} 
\usepackage{mathptmx} 
\usepackage{times} 
\usepackage{amsmath} 
\usepackage{amssymb}  
\usepackage{algorithm}
\usepackage{algpseudocode}
\usepackage{float} 
\usepackage[dvipsnames]{xcolor}
\usepackage{balance}
\usepackage{url}
\usepackage{cite}
\usepackage[hypertexnames=false]{hyperref}

\DeclareMathOperator*{\argmin}{arg\,min}

\title{\LARGE \bf
Learning Forward \& Reverse Skills from a Single Unfinished Demonstration for Constrained Manipulation Tasks
}

\author{Yexin Hu$^{1}$, Haoyi Zheng$^{1}$, Johannes Heidersberger$^{1}$ and Dongheui Lee$^{1, 2}$%
\thanks{$^{1}$Yexin Hu, Haoyi Zheng, Johannes Heidersberger and Dongheui Lee are with the Autonomous Systems Lab, Technische Universität Wien (TU Wien), Austria. Emails: \{yexin.hu, haoyi.zheng, johannes.heidersberger, dongheui.lee\}@tuwien.ac.at}%
\thanks{$^{2}$Dongheui Lee is with the Institute of Robotics and Mechatronics, German Aerospace Center (DLR), Germany.}%
\thanks{This work was partially supported by the European Union project INVERSE under grant agreement No.~101136067.}
}

\begin{document}

\maketitle
\thispagestyle{empty}
\pagestyle{empty}

\begin{abstract}

Learning from demonstration (LfD) enables robots to learn manipulation skills directly from expert demonstrations but remains challenging for contact-rich tasks involving geometric constraints and force interaction. Existing approaches typically require multiple complete demonstrations and do not support reverse skill execution. In this paper, we present a unified one-shot framework for constrained manipulation that learns both forward and reverse execution from a single, possibly unfinished demonstration. Our method decomposes demonstrations into non-contact and contact phases, with non-contact motion encoded with dynamic movement primitives (DMP), and contact motion represented as a sequence of screw motion primitives segmented by our proposed geometry-driven twist-direction segmentation algorithm. During execution, screw primitives are executed sequentially under admittance-guided pose correction and speed regulation, enabling task completion beyond the demonstrated trajectory length as well as reverse skill execution without additional learning data. Experiments on peg insertion, battery insertion, lock opening, and screw driving tasks demonstrate improved success rates and robustness over segmentation and one-shot trajectory learning baselines. Details are available on the project website: 
\href{https://tuwien-asl.github.io/LfD-Screw/}
{\nolinkurl{https://tuwien-asl.github.io/LfD-Screw/}}.

\end{abstract}

\section{Introduction}
Learning robot manipulation skills from human demonstrations is a promising paradigm that enables robots to operate in unstructured environments~\cite{ravichandar2020recent,correia2024survey}. However, many real-world tasks, such as insertion, fastening and unlocking, are inherently contact-rich and governed by geometric constraints. These tasks involve intermittent contact, changes in motion direction, and tight tolerances, which significantly challenge learning from demonstration methods.

Most existing LfD methods require multiple complete demonstrations to capture task variability. However, in contact-rich manipulation, such data can be expensive to collect, and demonstrations are frequently limited to single trial and may be unfinished due to safety constraints, physical limitation, or the repetitive sub-motions. Furthermore, skill learning is typically directional, preventing robots from learning reverse skills (e.g. learning locking from unlocking) without giving the additional demonstrations. These limitations motivate a one-shot framework that could learn from a potentially unfinished demonstration that supports both forward and reverse execution.

\begin{figure}[t]
  \centering
  \includegraphics[width=\columnwidth]{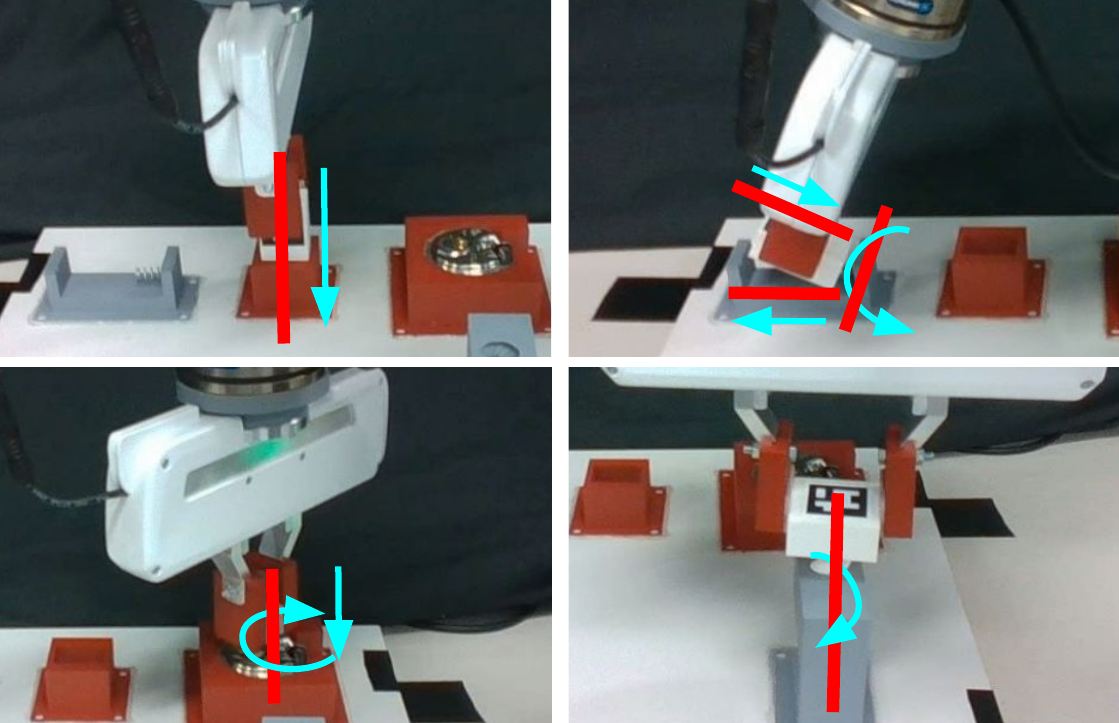}
  \vspace{-0.7cm}
  \caption{Tackled constrained manipulation tasks: peg insertion, battery insertion, lock opening and screw driving. Each task can be decomposed into sequences of screw motions, with red line indicating screw axis.}
  \label{fig:1}
\vspace{-0.7cm}
\end{figure}

However, enabling such one-shot learning in contact-rich settings is challenging due to its strong geometric constraints. While free-space motion can usually be effectively encoded using a trajectory-based method such as Dynamic Movement Primitives (DMP)~\cite{Fabisch2024, saveriano2023dynamic}, contact motion exhibits structured geometric behavior that is poorly captured alone by those methods. Classic control approaches address this through impedance or force-based control~\cite{hogan1984impedance, albu2003cartesian, mason2007compliance}, but typically require task-specific modeling, limiting their learning efficiency and generalizability across diverse tasks.

We observe that many contact-rich manipulation tasks can be naturally decomposed into sequences of constrained motions exhibiting consistent screw-like motion behavior, such as insertion, rotation and coupled translation-rotation (see Fig.~\ref{fig:1}). These geometric structures are invariant to the execution length and provide a principled basis for generalization beyond the demonstrated trajectory. 

Leveraging this insight, we propose a unified one-shot learning framework which decomposes the demonstration into non-contact and contact phases, with non-contact phase encoded with DMP; contact phase modeled as screw primitives~\cite{davidson2004robots, bahety2024screwmimic, mahalingam2022human} executed under wrench-aware control, enabling task completion learned from unfinished demonstration and supporting reverse execution without additional data. We validate the proposed framework on four representative contact-rich manipulation tasks: peg insertion, battery insertion, lock opening and screw driving as shown in Fig.~\ref{fig:1}. Across all tasks, our method achieves higher success rates compared to other one-shot trajectory-based baselines.

The contributions of this paper are:
\begin{itemize}
\item A unified one-shot learning framework for constrained contact-rich manipulation, enabling task completion and execution of both forward and reverse skills learned from a single demonstration.
\item A geometry-driven twist-direction trajectory segmentation algorithm for contact phase decomposition.
\item A screw-primitive representation with interaction-aware execution combining 6D admittance pose correction and 1D admittance progress control.
\end{itemize}

\section{Related Works}

\subsection{Learning from Demonstration for Manipulation}
Learning from demonstration has been widely studied for transferring manipulation skills from humans to robots~\cite{ravichandar2020recent, correia2024survey}. Recent vision-based and multimodal learning approaches ~\cite{chi2025diffusion, zeng2021transporter, zitkovich2023rt} have shown powerful performance, but typically rely on large-scale datasets or extensive pretraining. In contrast, kinesthetic representations such as Dynamic movement primitives (DMP) and Gaussian Mixture Models (GMM)~\cite{saveriano2023dynamic, Fabisch2024, ijspeert2013dynamical, calinon2010learning, calinon2016tutorial, pervez2018learning} enable smooth trajectory reproduction and generalization in free-space manipulation tasks learning from a single demonstration. However, these methods usually struggle in contact-rich tasks, where motion is governed by geometric constraints, and generally assume complete demonstrations without supporting reverse execution. Reverse execution has been studied at higher levels of abstraction, including reversible robot programs for error recovery~\cite{laursen2018modelling} and assembly-by-disassembly over part-level sequences~\cite{tian2024asap}, while reversibility of continuous skills learned from a single demonstration remains less explored.
\vspace{-0.1cm}
\subsection{Contact-Rich Manipulation and Force-Based Control}

Classical force and impedance control methods~\cite{hogan1984impedance, albu2003cartesian, mason2007compliance} enable robust physical interaction in constrained tasks such as peg-in-hole and surface following~\cite{song2016guidance, zhang2017force, van2018comparative, solanes2018adaptive, mohsin2019path}. However, these methods typically rely on task-specific constraint definitions or prior geometric knowledge~\cite{song2016guidance}, and do not directly infer the contact motion structure from demonstrations. As a result, they are not sufficient themselves alone for one-shot learning of general contact-rich skills.

\subsection{Screw Motion Primitives and Geometric Representations}
Screw theory~\cite{ball1998treatise, stramigioli2001geometry} provides compact representations of rigid-body motion suitable for most constrained tasks. However, existing screw-based approaches typically focus on free-space motion reproduction~\cite{verduyn2024enhancing, das2024screw, mahalingam2022human}, or representing the entire motion as a single global screw motion~\cite{bahety2024screwmimic, pettinger2022versatile}. Such formulations do not fully capture multi-phase contact-rich manipulation involving sequential geometric constraints.
\vspace{-0.1cm}
\subsection{Trajectory Segmentation and Skill Decomposition}
Trajectory-based segmentation decomposes demonstrations into reusable skills with kinematic heuristics or probabilistic models such as GMM~\cite{krishnan2017transition, shi2023waypoint, mahalingam2022human}. Segmentation methods with other modalities have also been applied widely to infer skill boundaries~\cite{zhang2024universal, potapov2014category, sliwowski2025m2r2}. However, these approaches are often sensitive to occlusions and discontinuous behaviors such as dwell, which do not mark true task transitions. In contact-rich manipulation, however, motion segments are often short, where kinematic noise can be amplified and speed changes are unreliable indicators of skill boundaries, such approaches are less reliable.

In summary, prior work reveals a gap in contact-rich manipulation learning. Trajectory-based LfD methods are data-efficient but lack explicit geometric constraints understanding; force controllers enable robust interaction but require task-specific knowledge; screw-based methods usually model free-space motion or single global primitives; and reversible-execution methods mainly operate at the robot-program or part-sequence level. Together with the difficulty of segmenting short constrained motions, these limitations motivate our geometry-driven, wrench-aware framework for one-shot contact-rich manipulation.

\section{Preliminaries: Screw Theory}

Screw theory provides a unified geometric framework for representing rigid-body motion in three-dimensional space. According to the Mozzi--Chasles theorem~\cite{lynch2017modern}, any instantaneous rigid-body motion can be represented as a rotation about, and a translation along, a single spatial line known as a screw axis. Such an instantaneous motion is represented by a twist $\boldsymbol{\xi}=(\boldsymbol{\omega}, \mathbf{v})\in \mathbb{R}^6$, where $\boldsymbol{\omega} \in \mathbb{R}^3$ denotes the angular velocity and $\mathbf{v} \in \mathbb{R}^3$ denotes the linear velocity.

The matrix representation of the twist in the Lie algebra $\mathfrak{se}(3)$ is given by
\begin{equation}
\hat{\boldsymbol{\xi}}=\begin{bmatrix}[\boldsymbol{\omega}]_\times & \mathbf{v}\\\mathbf{0}^\top & 0\end{bmatrix},
\end{equation}
where $[\boldsymbol{\omega}]_\times$ is the skew-symmetric matrix.

In the general case where $\boldsymbol{\omega} \neq \mathbf{0}$, the motion corresponds to a screw motion defined by a unit axis direction $\hat{\mathbf{s}}$, a point $\mathbf{q}$ on the axis, and a pitch $h \in \mathbb{R}$, which specifies the translation along the axis per unit rotation. Let $\dot{\theta}$ denote the rotational speed, the twist can be then written as
\begin{equation}
\label{twist_equation}
\boldsymbol{\xi} =\begin{bmatrix}\boldsymbol{\omega} \\
\mathbf{v}\end{bmatrix}
=\begin{bmatrix}\dot{\theta}\,\hat{\mathbf{s}} \\
-\,\dot{\theta}\,\hat{\mathbf{s}} \times \mathbf{q} + h\,\dot{\theta}\,\hat{\mathbf{s}}\end{bmatrix}.
\end{equation}

Pure translation is a special case with $\boldsymbol{\omega} = \mathbf{0}$, which can be interpreted as a screw motion with infinite pitch ($h \to \infty$) and no rotation. Then the twist reduces to $\boldsymbol{\xi}=(\mathbf{0}, \mathbf{v})$, and motion is fully modeled by the translation direction and magnitude.

Given a constant twist $\boldsymbol{\xi}$ and a scalar motion magnitude $\theta$, the corresponding finite rigid-body displacement is obtained through the exponential map:
\begin{equation}
\mathbf{T}(\theta) = \exp(\hat{\boldsymbol{\xi}}\,\theta) \in SE(3),
\end{equation}
which yields a rotation of angle $\theta$ about the screw axis together with a translation of magnitude $h\theta$ along the same axis. In the case of pure translation, the exponential map reduces to a translational displacement with no rotation.

\section{Method}

\subsection{Overview of the Learning and Execution Pipeline}
\begin{figure*}[t]
  \centering
  \includegraphics[width=\linewidth]{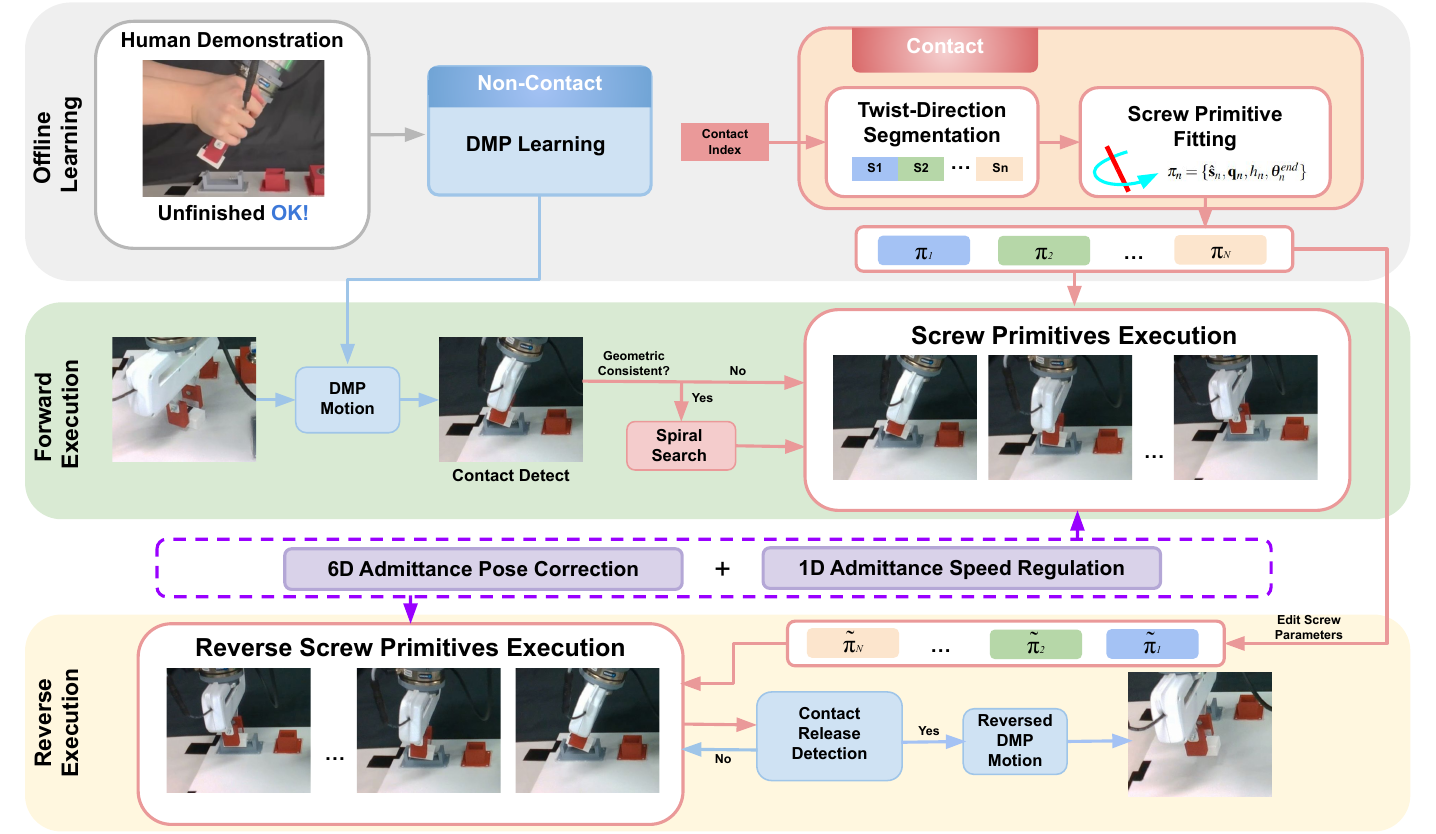}
  \vspace{-0.7cm}
  \caption{\textbf{Overview of the proposed framework.} A single forward demonstration is first encoded with a DMP. During execution, the DMP reproduces free-space motion until contact is detected. The contact phase is segmented using the proposed twist-direction method, and each segment is modeled as a screw primitive. Screw primitives are executed sequentially with 6D admittance-based pose correction and 1D admittance-based speed regulation. Reverse execution reuses the learned primitives with edited parameters and reversed order, followed by reversed DMP reproduction after contact release.}

  \label{pipeline}
\vspace{-0.7cm}
\end{figure*}

We consider a single human-guided demonstration consisting of an end-effector trajectory from the robot's initial grasp of the object until its final release. The proposed framework decomposes the demonstrated trajectory into non-contact and contact phases, and applies different learning and execution strategies to each phase (see Fig.~\ref{pipeline}).

The full demonstration trajectory is first encoded using a DMP, but is then only executed until contact is detected via wrench feedback. Then the demonstrated trajectory after this contact point is segmented using the proposed twist-direction-based method, which partitions motion based on the directional consistency in the twist space. Each segment is subsequently fitted with a screw motion primitive, generating a geometric representation of the constrained behavior.


At contact onset, the system checks whether local alignment is required by comparing short-window motion directions before and after the detected contact point in demonstration trajectory. If the directions remain consistent (i.e., high cosine similarity) while significant contact force is measured, the intended motion is considered blocked by misalignment. A basic sampled spiral search is then triggered after re-aligning the end-effector approach axis with the insertion direction while preserving rotation about that axis. This search is a common strategy in constrained tasks~\cite{van2018comparative} and the implement details are available in the website.


After that, the learned screw primitives are executed sequentially under wrench-aware admittance control, allowing contact-phase motion to extend beyond the demonstrated trajectory length when required. For an unfinished demonstration, each task-critical contact phase must be at least partially observed; the method extrapolates observed primitives but does not infer absent phases.

Our framework also supports reverse execution from the same demonstration. Learned screw primitives are reused, auto-edited, and executed in reverse order to follow the same geometric constraints in the opposite direction without relearning. After completing the reversed contact phases, the remaining free-space motion is executed with a DMP relearned from the time-reversed forward demonstration trajectory. Transitions from contact to free space are determined via wrench-based contact release detection.

\vspace{-0.1cm}
\subsection{Twist-direction-based Trajectory Segmentation}
We segment the contact-phase trajectory into a sequence of motion segments with constant twist directions. Unlike many segmentation methods that rely on time, velocity or dwell behavior, our segmentation uses only geometric motion directions, which is robust to short, noisy contact motions.

Prior to segmentation, the demonstrated pose trajectory is smoothed in $SE(3)$ by filtering incremental twists.
\subsubsection{\textbf{Direction-only twist features}} 
Given the smoothed trajectory $\{\mathbf{T}_i\}^{N-1}_{i=0} \in SE(3)$, we compute the per-step relative motion $\Delta \mathbf{T}_i = \mathbf{T}_i^{-1}\mathbf{T}_{i+1}$. And the corresponding twist is obtained using the logarithmic map $\hat{\boldsymbol{\xi}}_i \Delta \theta_i = \log(\Delta \mathbf{T}_i)$,
from which we extract the rotation axis direction $\mathbf{d}^\omega_i$ and translation direction $\mathbf{d}^v_i$, expressed in local frame, forming a 6D direction feature, $\mathbf{f}_i = ( \mathbf{d}_i^{\omega}, \mathbf{d}_i^{v} ) \in \mathbb{R}^6$. Negligible rotations and translations are treated as directionless and set to zero.

\subsubsection{\textbf{Block-wise dominant direction estimation}}


Given the per-step direction features $\mathbf{f}_i$, we partition the sequence into fixed-length blocks with length $B$. Within each block, we estimate the dominant rotation axis $\boldsymbol{\mu}^\omega$ and translation direction $\boldsymbol{\mu}^v$ separately by computing the principal direction of the unit direction samples with PCA. The sign ambiguity of PCA eigenvectors is resolved by aligning their direction with reference directions computed from the block endpoints. 

\subsubsection{\textbf{Block scanning and boundary detection}}


Blocks are scanned sequentially to form segments. For the current segment, dominant rotation and translation directions $(\mathbf{g}^\omega,\mathbf{g}^v)$ are estimated using the same procedure as for individual blocks. In each block $j$, the block-level directions
$(\boldsymbol{\mu}^\omega_j,\boldsymbol{\mu}^v_j)$ are compared to the current segment directions using cosine similarity $\langle\cdot\rangle$. A candidate boundary is detected if either the rotation direction or translation direction deviates beyond predefined thresholds $\tau_\omega$ and $\tau_v$:
$\langle \boldsymbol{\mu}^\omega_j,\mathbf{g}^\omega\rangle < \tau_\omega
 \lor \langle \boldsymbol{\mu}^v_j,\mathbf{g}^v\rangle < \tau_v$.
To suppress false positives caused by short-term fluctuations, we additionally require that the subsequent block $j+1$ should not remain consistent dominant rotational and translational direction with the current segment: $\neg(\langle \boldsymbol{\mu}^\omega_{j+1},\mathbf{g}^\omega\rangle > \tau_\omega \land 
\langle \boldsymbol{\mu}^v_{j+1},\mathbf{g}^v\rangle > \tau_v)$.
If both conditions hold, block $j$ is treated as a transition region; otherwise, it is absorbed into the current segment. 


\subsubsection{\textbf{In-block boundary refinement}}
\label{refinement}
When a transition is detected, we refine the changepoint within the transition block to obtain a sample-level boundary. Let $(\mathbf{g}^{\omega},\mathbf{g}^{v})$ denote the dominant directions of the current segment, and let $(\boldsymbol{\mu}^{\omega}_{j+1}, \boldsymbol{\mu}^{v}_{j+1})$ denote those of the subsequent block. For each candidate split index $k$ within the transition block, we minimize a direction-consistency cost that measures how well samples before $k$ align with the current segmentation direction, and how well samples after $k$ align with the new direction. The optimal boundary index $k^\star$ is obtained by:
\begin{equation}
k^\star = \argmin_{k \in \{0,\ldots,B-1\}}
\sum_{i<k} \ell(\mathbf{f}_i;\mathbf{g}^{\omega},\mathbf{g}^{v})
+
\sum_{i\ge k} \ell(\mathbf{f}_i;\boldsymbol{\mu}^{\omega}_{j+1},\boldsymbol{\mu}^{v}_{j+1}),
\end{equation}
where the per-sample cost $\ell(\cdot)$ is defined as
\begin{equation}
\ell(\mathbf{f}_i;\mathbf{a}^{\omega},\mathbf{a}^{v})
=
w_\omega \big(1 - \langle \mathbf{d}^\omega_i, \mathbf{a}^{\omega} \rangle \big)
+
w_v \big(1 - \langle \mathbf{d}^v_i, \mathbf{a}^{v} \rangle \big),
\end{equation}
with weights $w_\omega$ and $w_v$ balancing the contributions of rotation and translation and $\langle\cdot\rangle$ representing cosine similarity.

\subsubsection{\textbf{Global geometric refinement}}

Although block-level boundary detection produces locally consistent segments, short-term fluctuations in human demonstrations may introduce spurious boundaries. To enforce global geometric consistency, we compare each pair of adjacent segments by recomputing their dominant rotation and translation directions. If the cosine similarity of both rotational and translational directions exceeds the predefined same thresholds $\tau_\omega$ and $\tau_v$, the segments are merged. 

\subsubsection{\textbf{Short-motion (dwell) cleanup}}
\label{clean-up}

Human demonstrations often contain brief pauses or micro-corrections that generate segments with negligible net motion. Such segments do not represent meaningful contact-mode transitions. We therefore identify segments whose accumulated translation and rotation fall below predefined magnitude thresholds and merge them into adjacent segments. After this cleanup step, the segmentation algorithm outputs a set of segment indices $\{(s_n, e_n)\}_{n=1}^{N}$ that partition the trajectory and are subsequently used for screw fitting.

\subsection{Screw Primitive Learning}

\subsubsection{\textbf{Screw primitive parameterization}}
Given a contact-motion segment $\{\mathbf{T}_i\}^{N-1}_{i=0} \in SE(3)$ obtained from segmentation, we model the segment as a constant screw motion primitive. Let $\mathbf{T}_0$ denote the first pose of the segment and then the relative pose of each sample is $\mathbf{T}_i^{rel}=\mathbf{T}_0^{-1}\mathbf{T}_i$. The objective is to estimate a consistent twist $\boldsymbol{\xi}$ together with scalar progress variables $\theta_i$, such that:
\begin{equation}
\mathbf{T}_i^{rel} \approx \exp(\hat{\boldsymbol{\xi}}\,\theta_i).
\end{equation}

For the general screw case, we parameterize twist $\boldsymbol{\xi}$ using screw coordinates $(\hat{\mathbf{s}}, \mathbf{q}, h)$ as defined in (\ref{twist_equation}). Pure translation is treated as a special case described in sec.~\ref{pure_translation}.


\subsubsection{\textbf{Screw parameters optimization}}
We estimate a screw primitive by minimizing residuals in Lie algebra $\mathfrak{se}(3)$. For each sample $i$, the residual is defined as the logarithm of the pose discrepancy between the predicted screw transformation and the demonstrated relative pose, mapped to a 6D twist error in tangent space:
\begin{equation}
\mathbf{r}_i = \log\!\left( \exp\!\left( \hat{\boldsymbol{\xi}}\theta_i \right)^{-1} \mathbf{T}_i^{rel} \right) \in \mathbb{R}^6.
\end{equation}

Each residual is decomposed into rotational and translational components $\mathbf{r}_i^{\omega}$, $\mathbf{r}_i^{v}$, respectively. We solve the following weighted least-squares problem:
\begin{equation}
\min_{\hat{\mathbf{s}},\, \mathbf{q},\, h,\, \{\theta_i\}}
\sum_{i=0}^{N-1} \rho \! \left(\left\lVert \begin{bmatrix} w_R \, \mathbf{r}_i^{\omega} \\
w_T \, \mathbf{r}_i^{v} \end{bmatrix} \right\rVert_2\right),
\end{equation}
where $w_R$ and $w_T$ balance rotational and translational errors, and $\rho(\cdot)$ denotes Huber loss. Before optimization, each segment is resampled to achieve approximately uniform progress in $SE(3)$ space to reduce the sensitivity of the spatial imbalance in demonstration data. Compared with averaging per-step twists, this optimization enforces global $SE(3)$ consistency over the entire segment and jointly estimates the screw axis, pitch, and progress scale.

To ensure robustness and computational efficiency in one-shot setting, we adopt a reduced-parameter optimizing strategy. Instead of optimizing an independent progress variable $\theta_i$ for each sample, we initialize a monotone progress sequence $\theta_i^{*}$ with a global scale $s$ such that $\theta_i=s\theta_i^{*}$. This yields a low-dimensional optimization over $(\hat{\mathbf{s}}, \mathbf{q}, h, s)$, substantially improving the computational efficiency.

\subsubsection{\textbf{Pure translation segments}}
\label{pure_translation}
If the accumulated rotation within a segment is negligible, a direction $\hat{\mathbf{d}}$ is estimated with PCA and the twist is defined as $\boldsymbol{\omega}=\mathbf{0}, \mathbf{v}=\hat{\mathbf{d}}$. The progress $\theta_i$ corresponds to the displacement along the axis $\hat{\mathbf{d}}$. This case captures the insertion- or push-like geometric structures that are common in constrained tasks.

\subsubsection{\textbf{Primitive representation}}
Each segment is represented as a screw primitive $\pi_n=\{\hat{\mathbf{s}}_n, \mathbf{q}_n, h_n, \theta_n^{end}\}$. The terminal progress value $\theta_n^{end}$ is only used as a reference to initialize execution speed and is not used as a termination condition.

\subsection{Execution with Admittance-Guided Screw Primitives}
Each learned screw primitive $\pi_n=\{\hat{\mathbf{s}}_n, \mathbf{q}_n, h_n, \theta_n^{end}\}$ defines a constrained motion parameterized by the progress variable $\theta$ in the primitive's local frame. At each control cycle, a nominal pose is generated as 
\begin{equation}
    \mathbf{T}_{des}(\theta) = \mathbf{T}_0\exp(\hat{\boldsymbol{\xi}}_n\,\theta),
\end{equation}
where $\mathbf{T}_0$ is the initial robot end-effector pose at the start of primitive and $\hat{\boldsymbol{\xi}}_n$ is the twist associated with $\pi_n$. 

To enable robust interaction with the constrained environment, the nominal pose is modulated online with a 6D admittance correction defined in the end-effector (EE) frame. The small pose correction $\Delta\mathbf{x}_{ee}=[\Delta\mathbf{p}_{ee}, \Delta\boldsymbol{\theta}_{ee}]$ is generated by the measured wrench $\mathbf{w}_{ee} = [\mathbf{F}_{ee}, \boldsymbol{\tau}_{ee}]$ through standard admittance dynamics
\begin{equation}
    \mathbf{M}\Delta\ddot{\mathbf{x}}_{ee}  +  \mathbf{B}\Delta\dot{\mathbf{x}}_{ee}  +  \mathbf{K}\Delta\mathbf{x}_{ee}=\mathbf{w}_{ee}.
\end{equation}

In addition to pose correction, wrench feedback is also used to regulate execution speed along the screw. Before that, the screw axis direction $\hat{\mathbf{s}}_n$ and the point $\mathbf{q}_n$ on this axis, originally from local frame $\mathbf{T}_0$, are transformed into current EE frame as $\hat{\mathbf{s}}_{ee}$ and $\mathbf{q}_{ee}$. We decompose the measured wrench relative to the screw axis with axial components defined as
\begin{equation}
F_{\parallel} = \mathbf{F}_{ee}^{\top} \hat{\mathbf{s}}_{ee}, \;\;\;\; \tau_{\parallel} = \left( \boldsymbol{\tau}_{ee}
- \mathbf{q}_{ee} \times \mathbf{F}_{ee} \right)^{\top} \hat{\mathbf{s}}_{ee},
\end{equation}
and the perpendicular components defined as the magnitude of the residual wrench without their projections onto $\hat{\mathbf{s}}_{ee}$, denoted as $F_{\perp}$ and $\tau_{\perp}$. A scalar load is then computed as
\begin{equation}
L =
k_F^{\parallel} F_{\parallel}
+ k_T^{\parallel} \tau_{\parallel}
+ k_F^{\perp} F_{\perp}
+ k_T^{\perp} \tau_{\perp},
\label{eq:load}
\end{equation}
where $k_F^{\perp}, \,k_T^{\perp},\, k_F^{\parallel},\, k_T^{\parallel}$ are weight factors to balance each component. The scalar load $L$ is then used to regulate the progress velocity $\dot{\theta}$ through a 1D admittance model:
\begin{equation}
    M_{\theta}\dot{v}_{\theta} + B_{\theta}v_{\theta}=-L, \;\;\;\; \dot{\theta}=\dot{\theta}_0 + v_{\theta},
\end{equation}
where $\dot{\theta}_0$ is the nominal execution speed assigned based on $\theta_n^{end}$. As contact resistance increases, $\dot{\theta}$ is smoothly reduced. A screw primitive is considered complete when the effective progress velocity remains below a threshold for a sustained duration, after which the next primitive is executed until all the screw primitives are finished.

Here axial terms are assigned significantly larger weights, i.e., $k_F^{\perp}, k_T^{\perp} \ll k_F^{\parallel}, k_T^{\parallel}$, ensuring that speed regulation is primarily governed by resistance along the intended screw direction, while lateral interaction is mainly handled by 6D pose admittance.

\subsection{Reverse Skill Learning and Execution}
The proposed framework enables reverse task execution from a single forward demonstration. During forward learning, the contact phase is decomposed into an ordered sequence of screw primitives $\{\pi_n\}_{n=1}^{N}$. Reverse execution directly reuses these learned primitives to obtain $\tilde{\pi}_n$ by editing the learned screw parameters, i.e., $(\boldsymbol{\omega}, \mathbf{v}) \mapsto (-\boldsymbol{\omega}, -\mathbf{v})$. The reverse task is then executed by applying the sequence $\{\tilde{\pi}_N, \ldots, \tilde{\pi}_1\}$ under the same admittance control.

Reverse execution begins in contact. For each reversed primitive $\tilde{\pi}_n$, contact release is evaluated every short period using an active wrench-based probing test. Specifically, a small circular dither motion orthogonal to the screw axis is executed, and the load is monitored during this interval. The median absolute load is compared to a predefined threshold to determine whether the contact constraint has been released. Upon release, execution transitions to the next primitive or exits the contact phase. The remaining free-space motion is then reproduced using the DMP learned from the time-reversed trajectory of the forward demonstration.

\section{Experiments and Results}
\subsection{Experimental Setup}

All experiments are conducted on a Franka Emika Panda robot equipped with a wrist-mounted SCHUNK FT-AXIA80 force/torque sensor. Demonstrations are provided via kinesthetic teaching. For each task, only one single demonstration is used to learn both forward and reverse skills. During execution, only the end-effector pose and wrench measurements are used. AprilTags~\cite{olson2011apriltag} are used solely for initial object pose estimation; no visual feedback is used during task execution. Wrench safety limits are enforced to all tasks. 4 methods are evaluated on 5 randomized trials per task in both forward and reverse settings learned from both complete and unfinished demonstrations. With 4 tasks, this results in 80 trials per method in total. Segmentation parameters are fixed across all tasks; only safety threshold and scalar load weight in Eq.~\ref{eq:load} are task-dependent due to contact properties. Full values are listed on the website.

\begin{figure*}[t]
  \centering
  \includegraphics[width=\linewidth]{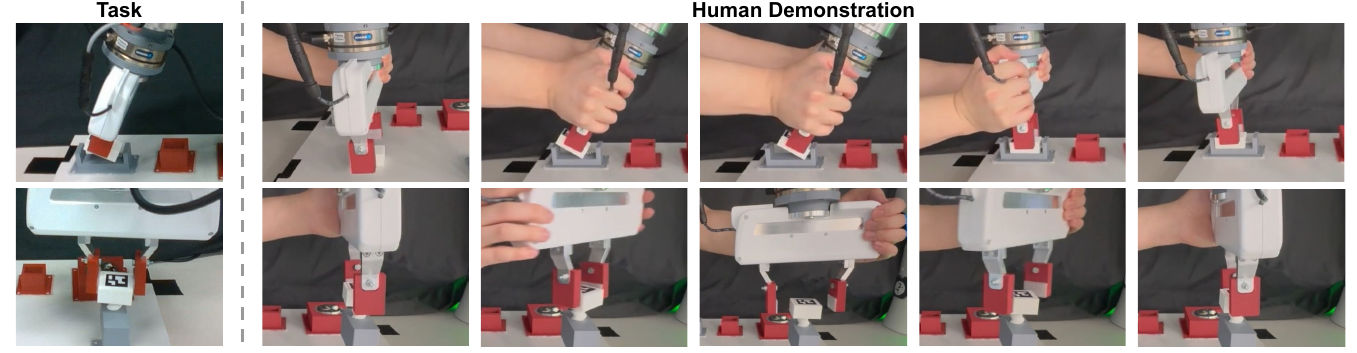}
  \vspace{-0.7cm}
  \caption{Expert demonstrations for the battery insertion and screw driving tasks. For battery insertion, key states along the full trajectory are shown. For screw driving, one representative cycle of grasp–rotate–release–reorient–regrasp is illustrated.}
  \label{demo}
\vspace{-0.7cm}
\end{figure*}

\vspace{-0.1cm}
\subsection{Tasks and Success Criteria}
\vspace{-0.1cm}
We evaluate four representative contact-rich manipulation tasks. Unfinished demonstrations are generated by directly truncating the contact phase of the full demonstration as shown in Fig.~\ref{fig:seg}:

\subsubsection{\textbf{Peg-in-hole insertion}} A rectangular peg is grasped and inserted into a fixed hole with approximately 1 mm dimensional clearance.  \textit{Success criteria}: (1) forward: insertion depth $>$ 2.3cm (max 2.5cm); (2) reverse: complete extraction without exceeding wrench limits.

\subsubsection{\textbf{Battery insertion}}  
The battery insertion involves three sequential contact phases as Fig.~\ref{demo}: (1) sliding along a spring channel; (2) tilting to contact the bottom surface; (3) final seating translation. 
\textit{Success criteria:} (1) forward: $\geq 75\%$ of the battery left surface lies inside the bed; (2) reverse: complete extraction.

\subsubsection{\textbf{Lock opening}}  
A pre-grasped key is inserted and rotated until fully opening the fixed lock with less than 0.5 mm clearance.  \textit{Success criteria:} (1) forward: lock fully opened; (2) reverse: complete key extraction.

\subsubsection{\textbf{Screw driving}}  
The screw is pre-inserted and partially threaded into the hole with approximately 1 mm clearance, and with two threads remaining before full tightening, and is then rotated until tight. During evaluation, trials are initialized with 1–3 threads remaining before full tightening. Execution begins directly in the contact phase.

Due to the wrist joint limits of the Panda robot, continuous rotation beyond a full revolution is not feasible without regrasping. During demonstration, human operator periodically releases, reorients and regrasps multiple times (see Fig.~\ref{demo}). To obtain a continuous $SE(3)$ trajectory suitable for screw primitive fitting, \textbf{grasped rotation segments} are stitched by chaining relative motions across regrasp intervals:
\begin{equation}
    \Delta{\mathbf{T}_k}=\mathbf{T}_0^{-1}\mathbf{T}_k,   \;\;\;\;   \mathbf{T}'_k=\mathbf{T}'_{last}\Delta{\mathbf{T}_k},
\end{equation}
where $\mathbf{T}_0$ is the first pose of the current grasp segment and $\mathbf{T}'_{last}$ is the last pose of the previous stitched segment, $\mathbf{T}_k$ and $\mathbf{T}'_k$ are the raw pose and the stitched pose within the segments at index $k$. The resulting stitched demonstration represents a single continuous screw-driving motion and is directly used for screw primitive fitting.

During execution, a similar regrasp-aware policy is adopted. We track the accumulated rotation about the learned screw axis. When the accumulated rotation exceeds approximately $90^\circ$, the robot (1) opens the gripper, (2) rotates the tool back by $90^\circ$ about the screw axis, (3) regrasps before continuing execution. This periodical execution enables arbitrary long screw rotations without violating the Panda robot's wrist joint limits.

\textit{Success criteria:} (1) forward: reach the screw tightening point, and stop by method's own mechanism; (2) reverse: back out the screw until final thread where the screw loses self-supporting stability and can no longer remain upright without external support.

\vspace{-0.1cm}
\subsection{Baselines}

We compare against three one-shot learning baselines to evaluate the contribution of geometric screw modeling and interaction-aware execution:

\subsubsection{\textbf{DMP only}}
The entire demonstrated trajectory (non-contact and contact), is reproduced using a single DMP. Contact detection, spiral search and wrench safety limits are identical to those used in our method.

\subsubsection{\textbf{DMP + admittance}}
The entire demonstrated trajectory is reproduced with a DMP under 6D Cartesian admittance correction, other settings are the same as DMP only.

\subsubsection{\textbf{DMP + twist + admittance}}
Non-contact phases apply DMP. Contact phases are segmented with our method, but each segment is represented by a constant twist computed from start–end displacement. Execution uses the same admittance correction and speed regulation as our method.

For reverse tasks, both our method and DMP + twist + admittance baseline assume that extraction must be achieved through contact-phase primitives; extraction achieved by DMP motion is not considered successful.

\subsection{Task Performance with Complete Demonstrations}



Table~\ref{tab:complete} reports task success rates for complete demonstrations. Our method achieves 20/20 success in both forward and reverse settings. DMP only performs well in insertion-dominant tasks but degrades in screw driving, where success depends on matching the demonstrated remaining thread length. DMP + Admittance improves some contact interaction but fails in highly constrained tasks such as lock opening and screw driving, where admittance-induced deviations can prevent the required geometric configurations to initialize next primitive. DMP + Twist + Admittance performs competitively in peg and battery tasks but fails in screw driving and reverse lock opening because start--end twists do not capture accumulated rotational structure or the correct reverse extraction geometry. These results highlight the importance of explicit screw primitive fitting for maintaining geometric consistency in constrained manipulation.

\begin{figure}[t]
  \centering
  \includegraphics[width=\linewidth]{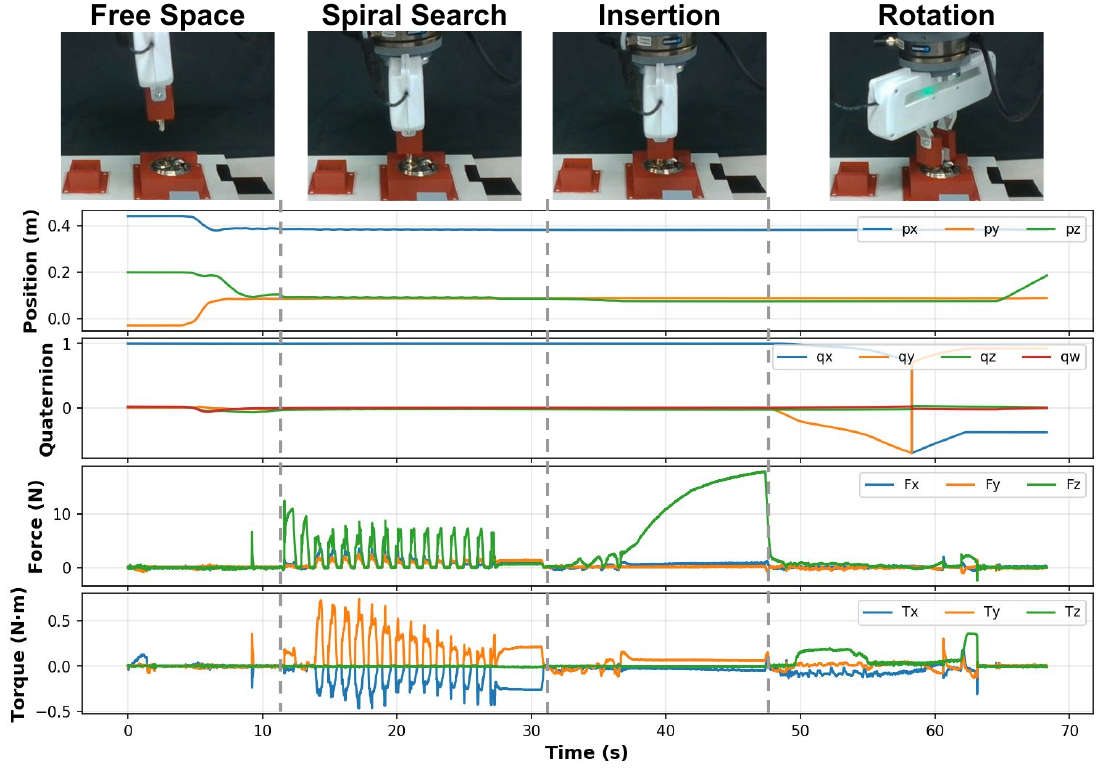}
  \vspace{-0.9cm}
  \caption{Each phase of lock opening task learned from unfinished demonstration, together with pose and wrench collected from execution.}
  \label{key_task}
\vspace{-0.2cm}
\end{figure}



\begin{figure*}[t]
  \centering
  \includegraphics[width=0.9\linewidth]{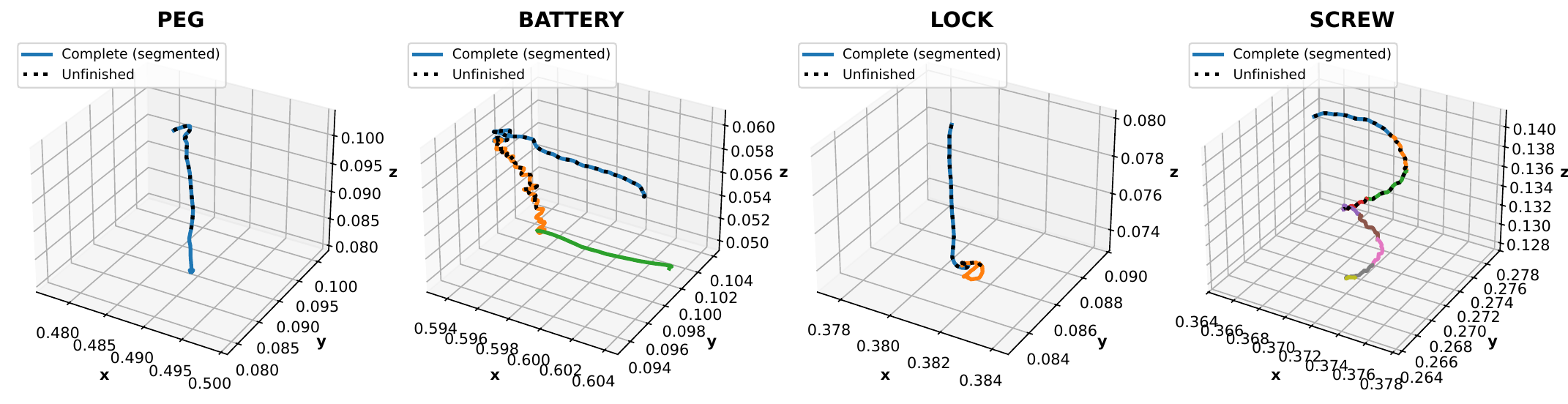}
  \vspace{-0.4cm}
  \caption{Segmentation of 4 tasks on complete demonstrations in contact phase with colored solid line, unfinished demonstration shown as black dot line.}
  \label{fig:seg}
\vspace{-0.5cm}
\end{figure*}


\begin{table}[t]
\centering
\caption{Success rate learned from \textbf{complete} demonstrations}
\vspace{-0.3cm}
\label{tab:complete}
\setlength{\tabcolsep}{2.5pt}
\renewcommand{\arraystretch}{1.05}
\begin{tabular}{l|cccc|c|cccc|c}
\hline
& \multicolumn{5}{c|}{\textbf{Forward}} & \multicolumn{5}{c}{\textbf{Reverse}} \\
\cline{2-11}
Method & Peg & Bat & Lock & Scr & Total & Peg & Bat & Lock & Scr & Total \\
\hline
\textbf{Ours} 
& 5/5 & 5/5 & 5/5 & 5/5 & \textbf{20/20} 
& 5/5 & 5/5 & 5/5 & 5/5 & \textbf{20/20} \\

DMP 
& 5/5 & 3/5 & 5/5 & 2/5 & 15/20 
& 5/5 & 5/5 & 4/5 & 0/5 & 14/20 \\

DMP+Adm 
& 5/5 & 5/5 & 0/5 & 2/5 & 12/20 
& 5/5 & 5/5 & 5/5 & 0/5 & 15/20 \\

DMP+Twist+Adm 
& 5/5 & 5/5 & 5/5 & 0/5 & 15/20 
& 5/5 & 5/5 & 0/5 & 0/5 & 10/20 \\
\hline
\end{tabular}
\vspace{-0.5cm}
\end{table}

\begin{figure}[t]
  \centering
  \includegraphics[width=\linewidth]{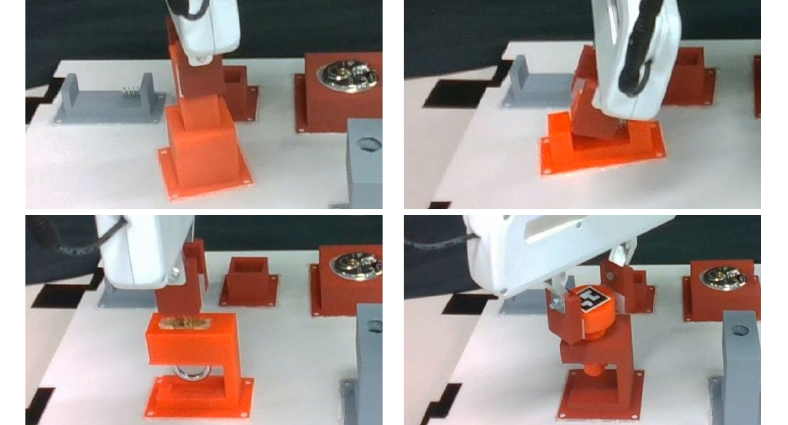}
  \vspace{-0.7cm}
  \caption{Generalization to novel objects of different dimension and target position for each task.}
  \label{new}
\vspace{-0.5cm}
\end{figure}

\subsection{Task Performance with Unfinished Demonstrations}

Table~\ref{tab:unfinished} evaluates truncated demonstrations. Fig.~\ref{key_task} shows lock opening learned from an unfinished demonstration, where wrench feedback regulates insertion and rotation progress through 1D admittance. DMP only and DMP + Admittance terminate at the demonstrated endpoint, lacking the mechanism to extend forward contact motion. DMP + Twist + Admittance extends motion but degrades in screw and reverse tasks due to the same geometric inconsistencies observed with complete demonstrations.

Our method achieves 18/20 success in both forward and reverse settings. Failures occur only in screw driving and are mainly caused by initial pose errors and pitch deviations from trajectory stitching, which accumulate over multiple rotations and lead to axis misalignment. Overall, results demonstrate that screw primitive modeling enables reliable task completion from partial demonstrations under geometric constraints.

\begin{table}[t]
\centering
\caption{Success rate learned from \textbf{unfinished} demonstrations}
\vspace{-0.3cm}
\label{tab:unfinished}
\setlength{\tabcolsep}{2.5pt}
\renewcommand{\arraystretch}{1.05}
\begin{tabular}{l|cccc|c|cccc|c}
\hline
& \multicolumn{5}{c|}{\textbf{Forward}} & \multicolumn{5}{c}{\textbf{Reverse}} \\
\cline{2-11}
Method & Peg & Bat & Lock & Scr & Total & Peg & Bat & Lock & Scr & Total \\
\hline
\textbf{Ours} 
& 5/5 & 5/5 & 5/5 & 3/5 & \textbf{18/20} 
& 5/5 & 5/5 & 5/5 & 3/5 & \textbf{18/20} \\

DMP 
& 0/5 & 0/5 & 0/5 & 2/5 & 2/20 
& 5/5 & 5/5 & 5/5 & 0/5 & 15/20 \\

DMP+Adm 
& 0/5 & 0/5 & 0/5 & 2/5 & 2/20 
& 5/5 & 5/5 & 5/5 & 0/5 & 15/20 \\

DMP+Twist+Adm 
& 5/5 & 5/5 & 4/5 & 0/5 & 14/20 
& 5/5 & 5/5 & 0/5 & 0/5 & 10/20 \\
\hline
\end{tabular}
\vspace{-0.4cm}
\end{table}

\vspace{-0.1cm}
\subsection{Segmentation Evaluation}

For segmentation evaluation, 5 demonstrations are collected for each task, resulting in 20 trajectories in total. Fig.~\ref{fig:seg} shows representative segmentation results for the four tasks used in execution experiments. For screw task, the segmentation is applied to each grasped rotation segment before stitching. 


We compare our twist-direction segmentation against Human-Guided Screw Planning (HGS)~\cite{mahalingam2022human}, Kernel Temporal Segmentation (KTS)~\cite{potapov2014category}, and Automatic Waypoint Extraction (AWE)~\cite{shi2023waypoint}, which are based on screw-motion reconstruction error, temporal signal variation, and waypoint reconstruction error, respectively.

Performance is evaluated using boundary F1 score~\cite{sliwowski2025m2r2} at tolerances $\tau \in {0.2, 0.5, 0.8s}$, Variation of Information (VI)~\cite{meilua2003comparing}, and segment count error (Cnt.). 
For fair comparison, all methods are applied with dwell clean-up to remove spurious micro-segments (see Sec.~\ref{clean-up}). 

Table~\ref{tab:seg_main} reports segmentation performance on four tasks. Our method achieves an overall better performance. 
 AWE relies primarily on waypoints reconstruction error, making it brittle in rotation-dominant scenarios. HGS incorporates screw-geometry reasoning and is therefore more aligned with geometric interpretation, achieving the lowest segment count error, but is less stable in boundary detection. KTS achieves the best performance among all three baselines, yet still underperforms compared to ours. In contrast, our method detects transitions based on twist-direction consistency in SE(3), considering both local directional changes and global twist coherence, yielding more stable boundaries for short and highly constrained motions.

\begin{table}[t]
\centering
\footnotesize
\caption{Segmentation performance on four tasks}
\vspace{-0.3cm}
\label{tab:seg_main}
\setlength{\tabcolsep}{3pt}
\renewcommand{\arraystretch}{1.1}
\begin{tabular}{lccccc}
\hline
Method & F1@.2 & F1@.5 & F1@.8 & VI & Cnt. \\
\hline
\textbf{Ours}
& \textbf{0.67$\pm$0.44}
& \textbf{0.92$\pm$0.24}
& \textbf{0.98$\pm$0.07}
& \textbf{0.29$\pm$0.32}
& 0.05$\pm$0.22 \\

HGS
& 0.6$\pm$0.48
& 0.75$\pm$0.41
& 0.91$\pm$0.25
& 0.39$\pm$0.43
& \textbf{0.00$\pm$0.00} \\

KTS
& 0.65$\pm$0.46
& 0.90$\pm$0.26
& 0.94$\pm$0.23
& 0.32$\pm$0.36
& 0.05$\pm$0.22 \\

AWE
& 0.56$\pm$0.38
& 0.67$\pm$0.31
& 0.79$\pm$0.23
& 0.72$\pm$0.56
& 2.50$\pm$3.68 \\
\hline
\end{tabular}
\vspace{-0.7cm}
\end{table}

\vspace{-0.1cm}
\subsection{Generalization to Novel Objects}
We further applied our learned screw primitives to a novel object with different dimension as shown in Fig~\ref{new} for each task. All four tasks are successfully completed with the screw primitives learned from previous same demonstration but only provided a target position under admittance control, which indicates that our proposed method captures task-consistent motion structure and exhibits transferability across objects with similar constraints.

\section{Conclusions}

In this paper, we present a unified one-shot learning framework for constrained, contact-rich manipulation that enables both forward and reverse task execution from a single, potentially unfinished demonstration. By decomposing motion into non-contact DMP reproduction and contact-phase screw primitives, the proposed approach captures geometric constraints while regulating execution through wrench-aware admittance control.

We introduce a twist-direction-based segmentation that enables task-agnostic decomposition of contact motion, from which screw primitives are learned to allow interaction-driven progress and termination beyond the demonstrated trajectory length. Experimental results across four real-world manipulation tasks demonstrate improved robustness and generalization compared to trajectory-based one-shot baselines.

Execution speed was not optimized and could be improved with faster alignment and primitive-speed selection. Remaining limitations include possible axis/pitch misalignment from sustained Cartesian admittance offsets during screw-driving motions with regrasping, and generalization mainly to objects with similar contact topology; future work will explore screw-aware admittance and online primitive adaptation.

Overall, this work highlights the benefits of integrating geometric screw representations with interaction-aware execution, providing a practical and data-efficient solution for constrained robot manipulation task learning.

\balance
\bibliographystyle{IEEEtran}
\bibliography{IEEEabrv,references}

@article{Fabisch2024,
  doi = {10.21105/joss.06695},
  url = {https://doi.org/10.21105/joss.06695},
  year = {2024},
  publisher = {The Open Journal},
  volume = {9},
  number = {97},
  pages = {6695},
  author = {Alexander Fabisch},
  title = {movement\_primitives: Imitation Learning of Cartesian Motion with Movement Primitives},
  journal = {Journal of Open Source Software}
}

@article{ravichandar2020recent,
  title={Recent advances in robot learning from demonstration},
  author={Ravichandar, Harish and Polydoros, Athanasios S and Chernova, Sonia and Billard, Aude},
  journal={Annual review of control, robotics, and autonomous systems},
  volume={3},
  number={1},
  pages={297--330},
  year={2020},
  publisher={Annual Reviews}
}

@article{correia2024survey,
  title={A survey of demonstration learning},
  author={Correia, Andre and Alexandre, Luis A},
  journal={Robotics and Autonomous Systems},
  volume={182},
  pages={104812},
  year={2024},
  publisher={Elsevier}
}

@article{saveriano2023dynamic,
  title={Dynamic movement primitives in robotics: A tutorial survey},
  author={Saveriano, Matteo and Abu-Dakka, Fares J and Kramberger, Alja{\v{z}} and Peternel, Luka},
  journal={The International Journal of Robotics Research},
  volume={42},
  number={13},
  pages={1133--1184},
  year={2023},
  publisher={SAGE Publications Sage UK: London, England}
}

@inproceedings{hogan1984impedance,
  title={Impedance control: An approach to manipulation},
  author={Hogan, Neville},
  booktitle={1984 American control conference},
  pages={304--313},
  year={1984},
  organization={IEEE}
}

@inproceedings{albu2003cartesian,
  title={Cartesian impedance control of redundant robots: Recent results with the DLR-light-weight-arms},
  author={Albu-Schaffer, Alin and Ott, Christian and Frese, Udo and Hirzinger, Gerd},
  booktitle={IEEE International conference on robotics and automation},
  volume={3},
  pages={3704--3709},
  year={2003},
}

@article{mahalingam2022human,
  title={Human-guided planning for complex manipulation tasks using the screw geometry of motion},
  author={Mahalingam, Dasharadhan and Chakraborty, Nilanjan},
  journal={arXiv preprint arXiv:2209.05672},
  year={2022}
}

@article{bahety2024screwmimic,
  title={Screwmimic: Bimanual imitation from human videos with screw space projection},
  author={Bahety, Arpit and Mandikal, Priyanka and Abbatematteo, Ben and Mart{\'\i}n-Mart{\'\i}n, Roberto},
  journal={arXiv preprint arXiv:2405.03666},
  year={2024}
}

@article{davidson2004robots,
  title={Robots and screw theory: applications of kinematics and statics to robotics},
  author={Davidson, Joseph K and Hunt, Kenneth H and Pennock, Gordon R},
  journal={J. Mech. Des.},
  volume={126},
  number={4},
  pages={763--764},
  year={2004}
}

@article{chi2025diffusion,
  title={Diffusion policy: Visuomotor policy learning via action diffusion},
  author={Chi, Cheng and Xu, Zhenjia and Feng, Siyuan and Cousineau, Eric and Du, Yilun and Burchfiel, Benjamin and Tedrake, Russ and Song, Shuran},
  journal={The International Journal of Robotics Research},
  volume={44},
  number={10-11},
  pages={1684--1704},
  year={2025},
  publisher={Sage Publications Sage UK: London, England}
}

@inproceedings{zeng2021transporter,
  title={Transporter networks: Rearranging the visual world for robotic manipulation},
  author={Zeng, Andy and Florence, Pete and Tompson, Jonathan and Welker, Stefan and Chien, Jonathan and Attarian, Maria and Armstrong, Travis and Krasin, Ivan and Duong, Dan and Sindhwani, Vikas and others},
  booktitle={Conference on Robot Learning},
  pages={726--747},
  year={2021},
  organization={PMLR}
}

@inproceedings{zitkovich2023rt,
  title={Rt-2: Vision-language-action models transfer web knowledge to robotic control},
  author={Zitkovich, Brianna and Yu, Tianhe and Xu, Sichun and Xu, Peng and Xiao, Ted and Xia, Fei and Wu, Jialin and Wohlhart, Paul and Welker, Stefan and Wahid, Ayzaan and others},
  booktitle={Conference on Robot Learning},
  pages={2165--2183},
  year={2023},
  organization={PMLR}
}

@article{ijspeert2013dynamical,
  title={Dynamical movement primitives: learning attractor models for motor behaviors},
  author={Ijspeert, Auke Jan and Nakanishi, Jun and Hoffmann, Heiko and Pastor, Peter and Schaal, Stefan},
  journal={Neural computation},
  volume={25},
  number={2},
  pages={328--373},
  year={2013},
  publisher={MIT Press One Rogers Street, Cambridge, MA 02142-1209, USA journals-info~…}
}

@article{calinon2010learning,
  title={Learning and reproduction of gestures by imitation},
  author={Calinon, Sylvain and D'halluin, Florent and Sauser, Eric L and Caldwell, Darwin G and Billard, Aude G},
  journal={IEEE Robotics \& Automation Magazine},
  volume={17},
  number={2},
  pages={44--54},
  year={2010},
  publisher={IEEE}
}

@article{calinon2016tutorial,
  title={A tutorial on task-parameterized movement learning and retrieval},
  author={Calinon, Sylvain},
  journal={Intelligent service robotics},
  volume={9},
  number={1},
  pages={1--29},
  year={2016},
  publisher={Springer}
}

@article{pervez2018learning,
  title={Learning task-parameterized dynamic movement primitives using mixture of GMMs},
  author={Pervez, Affan and Lee, Dongheui},
  journal={Intelligent Service Robotics},
  volume={11},
  number={1},
  pages={61--78},
  year={2018},
  publisher={Springer}
}

@article{song2016guidance,
  title={Guidance algorithm for complex-shape peg-in-hole strategy based on geometrical information and force control},
  author={Song, Hee-Chan and Kim, Young-Loul and Song, Jae-Bok},
  journal={Advanced Robotics},
  volume={30},
  number={8},
  pages={552--563},
  year={2016},
  publisher={Taylor \& Francis}
}

@article{zhang2017force,
  title={Force control for a rigid dual peg-in-hole assembly},
  author={Zhang, Kuangen and Shi, MinHui and Xu, Jing and Liu, Feng and Chen, Ken},
  journal={Assembly Automation},
  volume={37},
  number={2},
  pages={200--207},
  year={2017},
  publisher={Emerald Publishing Limited}
}

@article{van2018comparative,
  title={Comparative peg-in-hole testing of a force-based manipulation controlled robotic hand},
  author={Van Wyk, Karl and Culleton, Mark and Falco, Joe and Kelly, Kevin},
  journal={IEEE Transactions on Robotics},
  volume={34},
  number={2},
  pages={542--549},
  year={2018},
  publisher={IEEE}
}

@article{mason2007compliance,
  title={Compliance and force control for computer controlled manipulators},
  author={Mason, Matthew T},
  journal={IEEE Transactions on Systems, Man, and Cybernetics},
  volume={11},
  number={6},
  pages={418--432},
  year={2007},
  publisher={IEEE}
}

@article{solanes2018adaptive,
  title={Adaptive robust control and admittance control for contact-driven robotic surface conditioning},
  author={Solanes, J Ernesto and Gracia, Luis and Munoz-Benavent, Pau and Esparza, Alicia and Miro, Jaime Valls and Tornero, Josep},
  journal={Robotics and Computer-Integrated Manufacturing},
  volume={54},
  pages={115--132},
  year={2018},
  publisher={Elsevier}
}

@article{mohsin2019path,
  title={Path planning under force control in robotic polishing of the complex curved surfaces},
  author={Mohsin, Imran and He, Kai and Li, Zheng and Du, Ruxu},
  journal={Applied Sciences},
  volume={9},
  number={24},
  pages={5489},
  year={2019},
  publisher={MDPI}
}

@book{ball1998treatise,
  title={A Treatise on the Theory of Screws},
  author={Ball, Robert Stawell},
  year={1998},
  publisher={Cambridge university press}
}

@article{stramigioli2001geometry,
  title={Geometry and screw theory for robotics},
  author={Stramigioli, Stefano and Bruyninckx, Herman},
  year={2001}
}

@inproceedings{verduyn2024enhancing,
  title={Enhancing motion trajectory segmentation of rigid bodies using a novel screw-based trajectory-shape representation},
  author={Verduyn, Arno and Vochten, Maxim and De Schutter, Joris},
  booktitle={IEEE International Conference on Robotics and Automation (ICRA)},
  pages={7179--7185},
  year={2024},
}

@article{das2024screw,
  title={Screw Geometry Meets Bandits: Incremental Acquisition of Demonstrations to Generate Manipulation Plans},
  author={Das, Dibyendu and Patankar, Aditya and Chakraborty, Nilanjan and Ramakrishnan, CR and Ramakrishnan, IV},
  journal={arXiv preprint arXiv:2410.18275},
  year={2024}
}

@article{pettinger2022versatile,
  title={A versatile affordance modeling framework using screw primitives to increase autonomy during manipulation contact tasks},
  author={Pettinger, Adam and Alambeigi, Farshid and Pryor, Mitch},
  journal={IEEE Robotics and Automation Letters},
  volume={7},
  number={3},
  pages={7224--7231},
  year={2022},
  publisher={IEEE}
}

@article{krishnan2017transition,
  title={Transition state clustering: Unsupervised surgical trajectory segmentation for robot learning},
  author={Krishnan, Sanjay and Garg, Animesh and Patil, Sachin and Lea, Colin and Hager, Gregory and Abbeel, Pieter and Goldberg, Ken},
  journal={The International journal of robotics research},
  volume={36},
  number={13-14},
  pages={1595--1618},
  year={2017},
  publisher={SAGE Publications Sage UK: London, England}
}

@inproceedings{shi2023waypoint,
  title={Waypoint-Based Imitation Learning for Robotic Manipulation},
  author={Shi, Lucy Xiaoyang and Sharma, Archit and Zhao, Tony Z and Finn, Chelsea},
  booktitle={Conference on Robot Learning},
  pages={2195--2209},
  year={2023},
  organization={PMLR}
}

@inproceedings{zhang2024universal,
  title={Universal visual decomposer: Long-horizon manipulation made easy},
  author={Zhang, Zichen and Li, Yunshuang and Bastani, Osbert and Gupta, Abhishek and Jayaraman, Dinesh and Ma, Yecheng Jason and Weihs, Luca},
  booktitle={IEEE International Conference on Robotics and Automation (ICRA)},
  pages={6973--6980},
  year={2024},
}

@book{lynch2017modern,
  title={Modern robotics},
  author={Lynch, Kevin M and Park, Frank C},
  year={2017},
  publisher={Cambridge University Press}
}

@inproceedings{olson2011apriltag,
  title={AprilTag: A robust and flexible visual fiducial system},
  author={Olson, Edwin},
  booktitle={IEEE international conference on robotics and automation},
  pages={3400--3407},
  year={2011},
}

@inproceedings{potapov2014category,
  title={Category-specific video summarization},
  author={Potapov, Danila and Douze, Matthijs and Harchaoui, Zaid and Schmid, Cordelia},
  booktitle={European conference on computer vision},
  pages={540--555},
  year={2014},
  organization={Springer}
}

@article{sliwowski2025m2r2,
  title={M2R2: MultiModal Robotic Representation for Temporal Action Segmentation},
  author={Sliwowski, Daniel and Lee, Dongheui},
  journal={arXiv preprint arXiv:2504.18662},
  year={2025}
}

@inproceedings{meilua2003comparing,
  title={Comparing clusterings by the variation of information},
  author={Meil{\u{a}}, Marina},
  booktitle={Learning Theory and Kernel Machines: 16th Annual Conference on Learning Theory and 7th Kernel Workshop, COLT/Kernel, Washington, DC, USA, August 24-27. Proceedings},
  pages={173--187},
  year={2003},
  organization={Springer}
}

@article{laursen2018modelling,
  title={Modelling reversible execution of robotic assembly},
  author={Laursen, Johan Sund and Ellekilde, Lars-Peter and Schultz, Ulrik Pagh},
  journal={Robotica},
  volume={36},
  number={5},
  pages={625--654},
  year={2018},
  publisher={Cambridge University Press}
}

@inproceedings{tian2024asap,
  title={Asap: Automated sequence planning for complex robotic assembly with physical feasibility},
  author={Tian, Yunsheng and Willis, Karl DD and Al Omari, Bassel and Luo, Jieliang and Ma, Pingchuan and Li, Yichen and Javid, Farhad and Gu, Edward and Jacob, Joshua and Sueda, Shinjiro and others},
  booktitle={2024 IEEE International Conference on Robotics and Automation (ICRA)},
  pages={4380--4386},
  year={2024},
  organization={IEEE}
}

\end{document}